\documentclass{article}

\usepackage{microtype}
\usepackage{graphicx}
\usepackage{subfigure}
\usepackage{booktabs} 

\usepackage{hyperref}

\usepackage[accepted]{mlsys2022}

\mlsystitlerunning{AutoNLU: Detecting, root-causing, and fixing NLU model errors}

\begin{document}

\twocolumn[
\mlsystitle{AutoNLU: Detecting, root-causing, and fixing NLU model errors}



\mlsyssetsymbol{equal}{*}

\begin{mlsysauthorlist}
\mlsysauthor{Pooja Sethi}{fb,stanford}
\mlsysauthor{Denis Savenkov}{fb}
\mlsysauthor{Forough Arabshahi}{fb}
\mlsysauthor{Jack Goetz}{fb}
\mlsysauthor{Micaela Tolliver}{fb}
\mlsysauthor{Nicolas Scheffer}{fb}
\mlsysauthor{Ilknur Kabul}{fb}
\mlsysauthor{Yue Liu}{fb}
\mlsysauthor{Ahmed Aly}{fb}
\end{mlsysauthorlist}

\mlsysaffiliation{fb}{Facebook, Redmond, Washington, USA}
\mlsysaffiliation{stanford}{Stanford University, Stanford, California, USA}

\mlsyscorrespondingauthor{Pooja Sethi}{pjasethi@stanford.edu}
\mlsyscorrespondingauthor{Denis Savenkov}{denxx@fb.com}
\mlsyscorrespondingauthor{Yue Liu}{yuei@fb.com}

\mlsyskeywords{Machine Learning, MLSys, Natural Language Understanding, Active Learning}

\vskip 0.3in

\begin{abstract}
Improving the quality of Natural Language Understanding (NLU) models, and more specifically, task-oriented semantic parsing models, in production is a cumbersome task. In this work, we present a system called AutoNLU, which we designed to scale the NLU quality improvement process. It adds automation to three key steps: detection, attribution, and correction of model errors, i.e., bugs. We detected four times more failed tasks than with random sampling, finding that even a simple active learning sampling method on an uncalibrated model is surprisingly effective for this purpose. The AutoNLU tool empowered linguists to fix ten times more semantic parsing bugs than with prior manual processes, auto-correcting 65\% of all identified bugs.
\end{abstract}
]


\printAffiliationsAndNotice{}  

\section{Introduction}

Natural Language Understanding (NLU) models are widely used by virtual assistants to complete task-oriented requests. For example, given a request in natural language like, \textit{``Play my running playlist,"} an NLU model returns a structured representation that could be more easily executed downstream. Typically, NLU models classify a user's request into a \textit{domain}, and further parse the request into \textit{intents} and \textit{slots} \cite{atis, gupta}. \footnote{NLU can include other NLP tasks such as disambiguation, coreference resolution, etc. For the purposes of this paper, we use NLU to refer to task-oriented semantic parsing.}

Developers, or linguists, frequently desire to add new domains, or to expand the coverage of existing domains to support a wider variety of user requests. Unfortunately, ensuring the quality of new domains is high or continuously improving existing domains is non-trivial. Each step of the quality improvement process, from discovering areas where the model is failing, understanding why it is failing, and finally, applying a fix, is highly manual.

We first look at discovering model bugs. Traditional approaches to discovering model bugs are either too fine or too coarse-grained. The traditional software engineering quality assurance (QA) approach is too fine-grained. In the QA approach, the model is hand-tested one-by-one, either with utterances that are construed to be prototypical of that domain or adversarial examples. The traditional ML approach is too coarse-grained. In the ML approach, a model is evaluated on a randomly-sampled validation set and test set; metrics on these sets may obfuscate the presence of bugs in a relatively small domain. Slicing, as discussed by \cite{checklist}, can be a great way of finding and inspecting bugs that would otherwise be hidden. However, there is still an opportunity cost in annotating a large validation and test set where bugs may be relatively rare. Active learning is often used to build a training set in a cost-effective way \cite{hitl}; however, it is not typically framed with the explicit goal of finding bugs. 

Once bugs are detected, developers hope to correct them. However, before they can correct the bug, they must understand what is causing them. Unfortunately, this is also non-trivial. Model interpretability tools can help, but fundamentally bugs often stem from training data \cite{data_validation}, which can be time-consuming to explore.

Finally, the quality improvement process is slowed down by a lack of tools to quickly fix bugs. While a machine learning engineer's instinct may be to try a new architecture, pre-training, or a parameter sweep, a domain confusion could be caused by lack of training data or even a mislabel. Linguists often have to create \textit{data patches}, i.e., data augmentations or relabels, by hand without tools to assist them.

These three challenges significantly slow down iteration speed, and we propose a more scalable approach to finding and fixing NLU bugs. In this work:
\begin{itemize}
\item We describe and demonstrate a system tailored to scaling the quality improvement process for task-oriented NLU models (Sections \ref{section:autonlu}, \ref{section:ui}). 
\item We present a fresh take on the objective of active learning for NLU, using it to continuously detect bugs as opposed to improve sample efficiency. We found that even a simple uncertainty sampling method is four times more effective than random sampling at finding utterances where the model will make a mistake (Sections \ref{section:detection}, \ref{section:simple_uncertainty_eval}).
\item We describe an approach to automatic error attribution and data augmentation and show that they are effective ways of fixing NLU bugs in large volumes (Sections \ref{section:error_classes}, \ref{section:correction}, \ref{section:data_aug_eval}). 
\end{itemize}

\section{AutoNLU}
\label{section:autonlu}

We designed a system called AutoNLU, which ties together and provides automation around three key steps to quality improvement.

\paragraph{Detection} AutoNLU re-frames the goal of active learning to maximizing \textit{bug detection}: find all observable examples that may be parsed incorrectly, rank them by importance (e.g., frequency), and annotate as many as possible within the available annotation budget. We are particularly interested in correcting ``known unknowns" \cite{hitl}. Traditional active learning setups, on the other hand, aim to increase \emph{sample efficiency}: annotate the fewest possible examples to reach acceptable performance on a test set. The flaw of this framing is that it assumes the test distribution is static. In contrast, in production settings, the test distribution is dynamic and thus it's desirable to make model improvements continuously. Moreover, sample efficiency is often not a concern by the time the model is put in production -- models are bootstrapped with pre-labeled crowd-sourced or generated data. 

\paragraph{Error Attribution} Once we have obtained a collection of bugs, AutoNLU attempts to attribute their root cause. We provide a small set of high-level error classifications that map directly to actionable next steps that a developer can take to fix the bug. These classifications can be made more specific and accurate over time. 

\paragraph{Correction} Finally, based off the error classification from the previous step, AutoNLU suggests an appropriate fix. For example, if the error is that there is not enough training data similar to the bug utterance, AutoNLU generates similar utterances that could be added to the training set. 
\subsection{Defining an NLU Bug}
Using the conventions described by \cite{gupta}, given a user request like, \textit{``Play my running playlist,"} our NLU model should predict it belongs to the \textit{music} domain, and produce a parse like \break \texttt{[IN:PLAY\_MUSIC Play my [SL:PLAYLIST\_NAME running] [SL:MUSIC\_TYPE playlist]]}, 

where \texttt{IN:} is the prefix of the intent, and \texttt{SL:} is the prefix of all the slots within the intent.\footnote{An analogy to an intent is a function or API call. Similarly, and analogy to a slot is a function argument.} If the domain or intent-level classification of the sentence is incorrect, or if an expected slot is missing or has the incorrect span, or if an extra slot is produced, we consider the prediction incorrect and the corresponding input a \textit{bug}. For example, for a user request like, \textit{``Play my holiday cooking playlist,"} if the NLU model produced a parse like \break \texttt{[IN:PLAY\_MUSIC Play my holiday cooking [SL:MUSIC\_TYPE playlist]]}, 
we would consider the request a bug because the \texttt{SL:PLAYLIST\_NAME} slot is missing.

\subsection{NLU Model Architecture}
Throughout this paper, we loosely use the term NLU model to actually describe a collection of two sets of models. The first set consists a single domain classifier. The second set contains many sequence-to-sequence parsing models, which jointly produce the intent and slot predictions. There is a parser per domain (or per small grouping of similar domains). The domain classifier routes the request to the appropriate intent-slot parser. 

The domain classifier is a BiLSTM model with an MLP output layer \cite{bilstm_crf}. The intent-slot parsers have a RoBERTA \cite{roberta} encoder and have either a non-autoregressive (NAR) decoder, similar to that described by \cite{span_pointer}, or an autoregressive BERT decoder \cite{bert}.

\section{System Design}
\begin{figure}[t]
  \includegraphics[scale=.2]{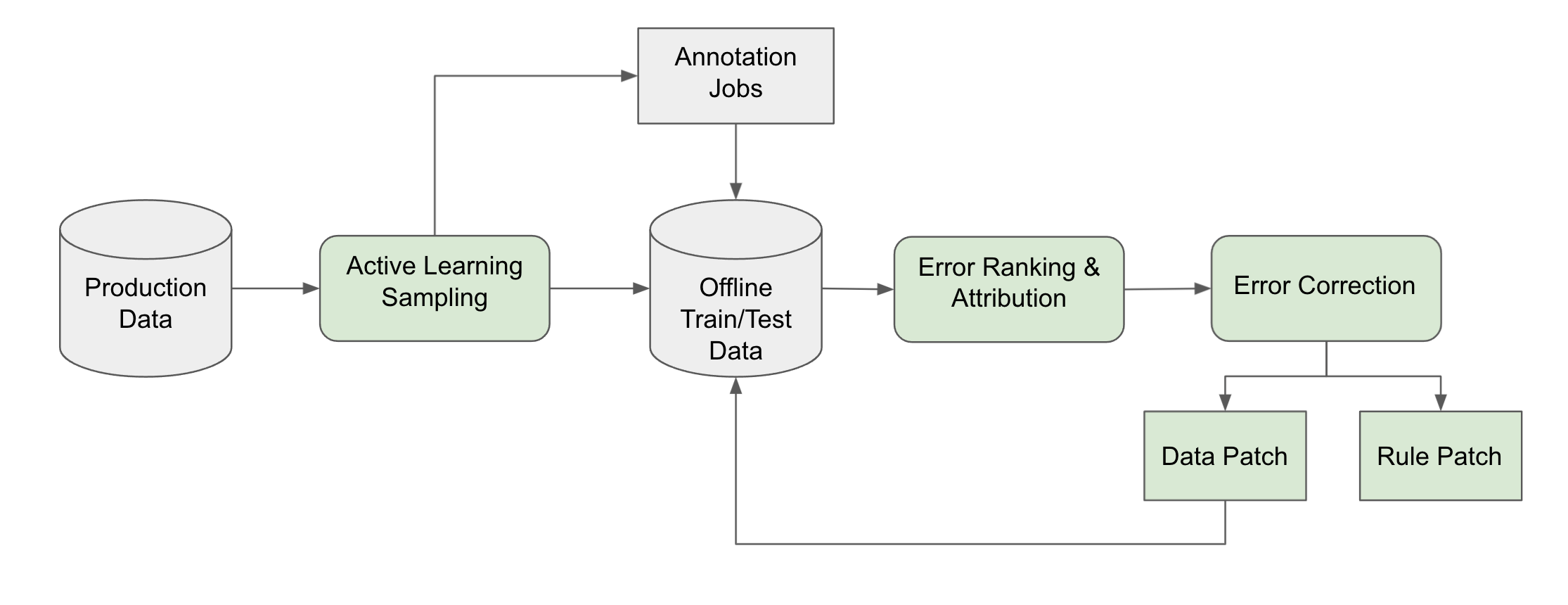}
  \caption{This diagram illustrates the flow of production data. The components highlighted in green are discussed in more detail in the following sections.}
  \label{fig:architecture}
\end{figure}

The high-level system design of AutoNLU is shown in Figure \ref{fig:architecture}.

\subsection{User Interface}
\label{section:ui}
\begin{figure*}[t]
  \centering
  \includegraphics[scale=.3]{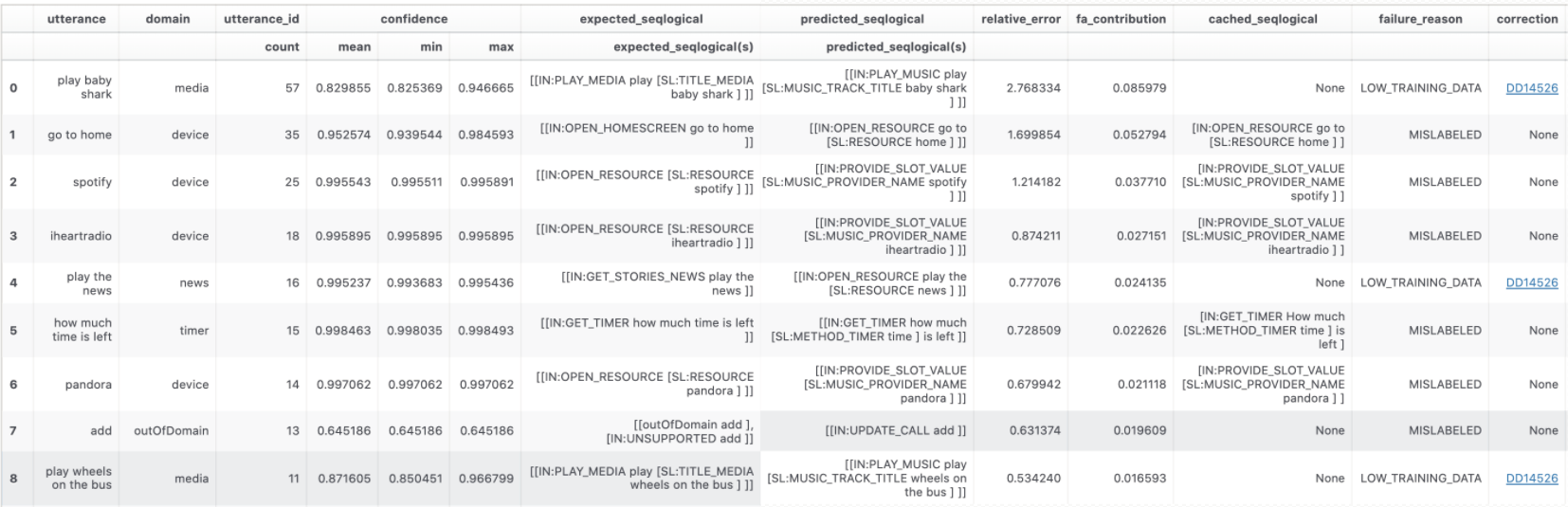}
  \caption{AutoNLU produces results as a pandas DataFrame \cite{pandas_paper, pandas_software}, viewable from a Jupyter notebook \cite{jupyter} . It displays confidence scores, the expected and predicted parse, the estimated contribution of the bug to overall accuracy, the failure reason, and where possible, an auto-generated correction. It also displays additional debugging information such as the training set annotation, under \texttt{cached\_seqlogical}.}
  \label{fig:pandas_interface}
\end{figure*}

\begin{figure}[t]
  \centering
  \includegraphics[scale=.09]{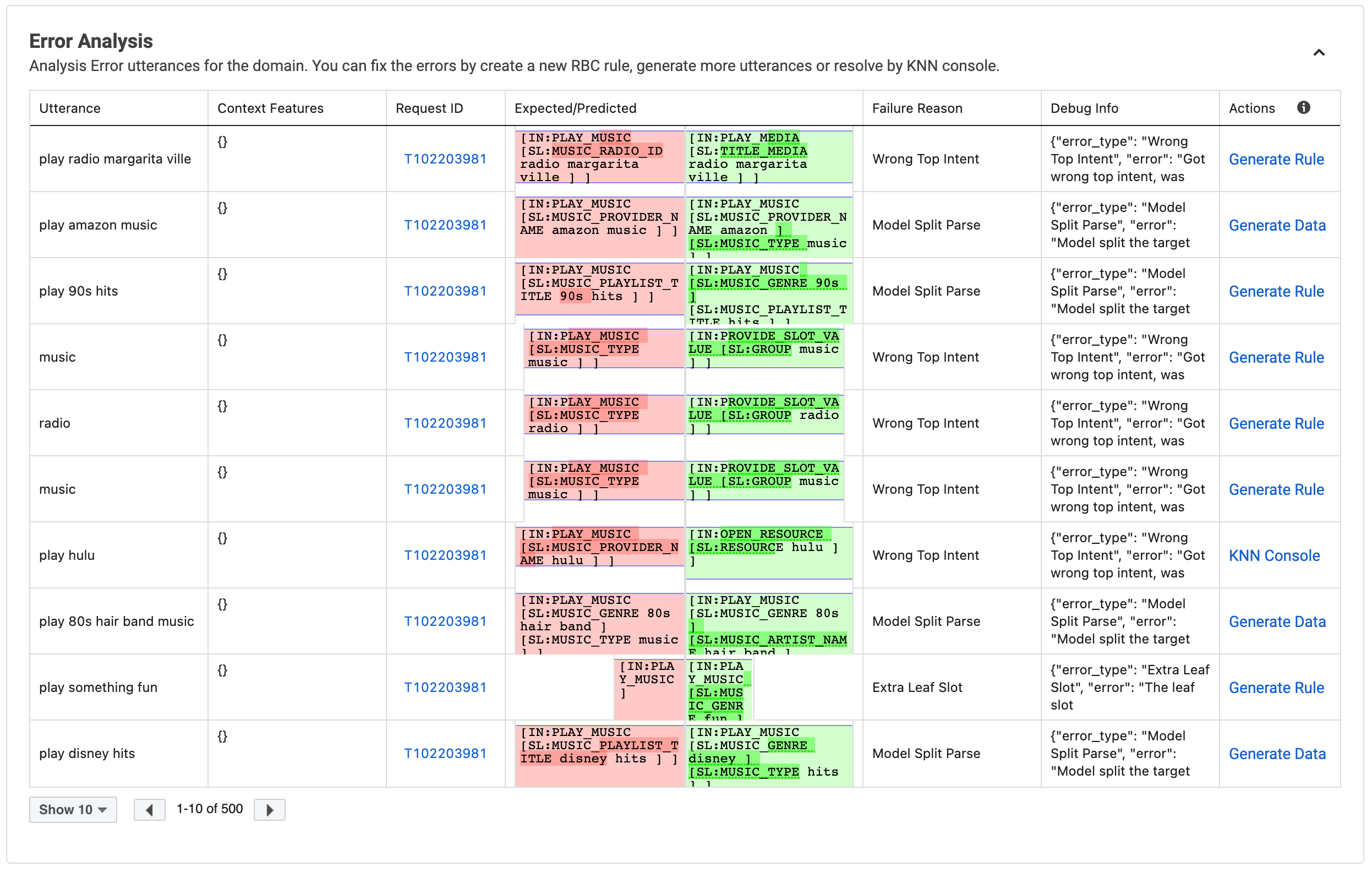}
  \caption{This is a screenshot that illustrates the web app version of the user interface for AutoNLU. It includes diff highlighting between the expected and predicted parse. Note that the data shown here is dummy data.}
  \label{fig:pretty_interface}
\end{figure}

The user interface (UI) for AutoNLU is a table showing a list of bugs for a domain developer or linguist to fix. By default, the bugs are ranked by their frequency. However, the ranking could also take into account other factors such as recency or uncertainty score. The UI also shows the user the failure reason and a suggested action to fix the bug. Examples are shown in Figures \ref{fig:pandas_interface} and \ref{fig:pretty_interface}.

\subsection{Detection}
\label{Detection}
\label{section:detection}
When designing our algorithm for bug detection, we kept the following principles in mind:
\begin{itemize}
    \item It should identify utterances that the NLU model will parse incorrectly.
    \item It should provide a numerical score that can allow us to rank and compare unlabeled production data (candidates) for sampling.
    \item Ideally, it should be feasible to run on hundreds-of-thousands to millions of examples. Else, we would have to decrease the size of our candidate set.
\end{itemize}

\subsubsection{Simple Uncertainty Baseline}
\label{section:simple_uncertainty}
The algorithm we use for bug detection is least-confidence uncertainty sampling, similar to that described by \cite{hitl}. This method assigns high \textit{uncertainty} scores for candidates where the model prediction had low \textit{confidence}.

For all \textit{n} candidates, we assign an uncertainty score of \texttt{1 - intent\_confidence}. We then select the top \textit{k} samples for human grading according to this score.

For simplicity, we do not use domain and slot confidence scores when computing the uncertainty score. If the semantic frame contained nested intents, we only use the score from the top-level intent. Combining uncertainty scores across the semantic frame is an interesting future direction.

\subsubsection{Intrinsic vs. Extrinsic Sampling Methods}
One of the key considerations for the detection algorithm was whether we should use an \textit{intrinsic} or \textit{extrinsic} approach. By intrinsic, we mean using the model confidence, scores, or logits to decide what to sample. By extrinsic, we mean training a meta-model that can predict when the base, i.e., NLU model will make a mistake \cite{safe_predict}. 

An advantage of the intrinsic approach is that, unlike extrinsic, it does not require training and maintaining a separate model. However, the downside is that it may rely on the base model being well-calibrated. Deep learning models are known to produce over-confident predictions even when they are incorrect \cite{overconfidence}. 

An advantage of the extrinsic approach is that it may allow you to consider other features that are available in logs but not at inference (e.g., Did the user repeat their request? -- which could indicate frustration from an incorrect NLU prediction). 

We went with the intrinsic, least-confidence uncertainty method because it was simple to implement and computationally inexpensive, since we already stored the logged model prediction and confidence scores from run-time on our candidates. Despite the lack of calibration, we found we were still able to get good results as described in Section \ref{section:simple_uncertainty_eval}. 

\subsection{Error Attribution}
\label{section:error_classes}
Once we've sampled bugs, the next step is to identify their root cause. The key principle we held in mind for error attribution is that the error categories must be actionable, i.e., map to a concrete next step that can be applied automatically or by a linguist to fix the bug. This mapping is shown in Table \ref{table:error_types} and the corrections are described in Section \ref{section:correction}.

We determine the error attribution by sequentially checking the cases below, in the order provided; the attribution falls through to \textit{unknown} if none of the criteria are met. Note that we use the term \textit{golden annotation} to refer to the annotation given to the bug utterance.

\subsubsection{Rule Mismatch}
In addition to using the NLU model to parse language requests, we have a small set of high-precision rules that trigger on must-pass requests. Sometimes the parse produced by a rule conflicts with the golden annotation provided for the bug. In this case, either the rule or the golden annotation needs to be updated so that they are in agreement.

\subsubsection{Mislabeled}
A surprisingly common cause of bugs is a labeling error, either in the training data or of the golden annotation itself. We assign this attribution whenever we see an exact match of the bug utterance in the training set, the annotation in training conflicts with the golden annotation, and the model's confidence in its prediction is above some threshold $\lambda$. We set $\lambda >= 0.9$. These signals indicate that the golden annotation may be incorrect.

A golden annotation can be incorrect for many reasons. First, labeling bugs is a more cognitively difficult task than labeling training data. Bugs are not pre-sorted by domain, and by nature of being bugs, may be more linguistically challenging to annotate than typical training data. Second, the people who create training data and label golden data are usually not the same. Training data is created under the instruction of linguists, who are subject-matter experts. Golden annotation of incoming production data is often done with the help of third-party vendors. And third, annotation guidelines can be difficult to keep up-to-date. Often the \textit{ontology} (label set) is updated faster than annotation guidelines can keep up.

\subsubsection{Low Training Data}
If a bug has made it past the first two filters, there's a good sign that there is a real issue with the model. Often bugs arise because there is not enough training data similar to the bug in order for the model to generalize. For the example used earlier, \texttt{[IN:PLAY\_MUSIC Play my holiday cooking [SL:MUSIC\_TYPE playlist]]}, the model may have not seen \textit{``holiday cooking,"} or something similar being used as playlist name. 

We have a simple heuristic to check if a bug should get the low training data error attribution: we check for an exact match of the bug utterance in the training data, and if none exists, we assign this error. This heuristic could over-trigger, and in the future we hope to change it to look for \textit{similar} training utterances as opposed to exact matches. For example, if the training set did not have an exact match but had the utterance, \textit{``Play my Christmas baking playlist,"} then we should not assign the low training data attribution, since this example should have been sufficient for the model to generalize.

\subsubsection{Unknown}
Finally, if none of the above criteria pass, we assign the bug to the unknown category.

\subsection{Correction}
\label{section:correction}

Once the error attribution is known, the appropriate correction can be applied. We discuss these potential corrections in further detail.

\begin{table}[t]
\centering
\caption{Each error attribution maps to an appropriate correction.}
\label{table:error_types}
\vskip 0.15in
\begin{small}
\begin{sc}
\begin{tabular}{lcccr}
\toprule
Error Attribution & Appropriate Correction \\
\midrule
Rule Mismatch        & Fix Rule\\
Mislabeled           & Fix Annotation Conflicts\\
Low Training Data    & Generate Data\\
Unknown              & Generate Rule\\
\bottomrule
\end{tabular}
\end{sc}
\end{small}
\vskip -0.1in
\end{table}

\subsubsection{Data Augmentation}
\label{section:data_aug_tool}
In the low training data scenario, AutoNLU automatically applies templated data augmentation with slot-based replacement. For example, given the bug utterance, \textit{``Play my holiday cooking playlist,"} it may produce the template \texttt{Play my $<$SL:PLAYLIST\_NAME$>$ playlist}. The template can then be expanded to new training examples, such as \textit{``Play my baking playlist,"} \textit{``Play my road trip playlist,"} etc.

While we used this templated approach for our experiments to have more control of the generated data, one could conceivably use language models for text generation instead.

\subsubsection{Annotation Conflict Correction}
In the case where there is a conflict between the golden annotation of the bug and the training data, AutoNLU shows the user the conflicting annotations. The user can then manually update the annotations, or use a labeling library for semantic frames, which we call \textit{data transformers}, to update them. The data transformer library supports operations such as renaming or adding intents and slots on large batches of data.

\subsubsection{Rule Generation}
In the rule mismatch case, we may need to update an incorrect rule. In the case where the error attribution is unknown, and it's not possible to manually inspect and find the root cause of the bug, our only option may be to generate a new high-precision rule, 

Rule generation is simple to do given the golden annotation and utterance as inputs. For our previous example, we could create the rule \texttt{IN:PLAY\_MUSIC$<$Play my $<$SL:PLAYLIST\_NAME holiday cooking$>$ $<$SL:MUSIC\_TYPE playlist$>>$} which would trigger if the same utterance was received at runtime. The downside of adding a rule as opposed to training data for the NLU model is, of course, limited generalization.

\section{Experiments}

Our evaluation of the AutoNLU system focused on its ability to detect bugs and to auto-correct them. We found that even our simple methods for sampling and data generation led to significant increases in the number of bugs detected and fixed.

\subsection{Simple Uncertainty}
\label{section:simple_uncertainty_eval}

Our evaluation of uncertainty sampling was split into two stages, an offline and an online evaluation. In the offline evaluation, we aimed to understand the relationship between NLU misclassifications and the uncertainty score. In the online evaluation, we aimed to understand the effectiveness of uncertainty sampling at finding failed tasks on production data.

\subsubsection{Offline Evaluation}
\begin{figure}[t]
  \centering
  \includegraphics[scale=.15]{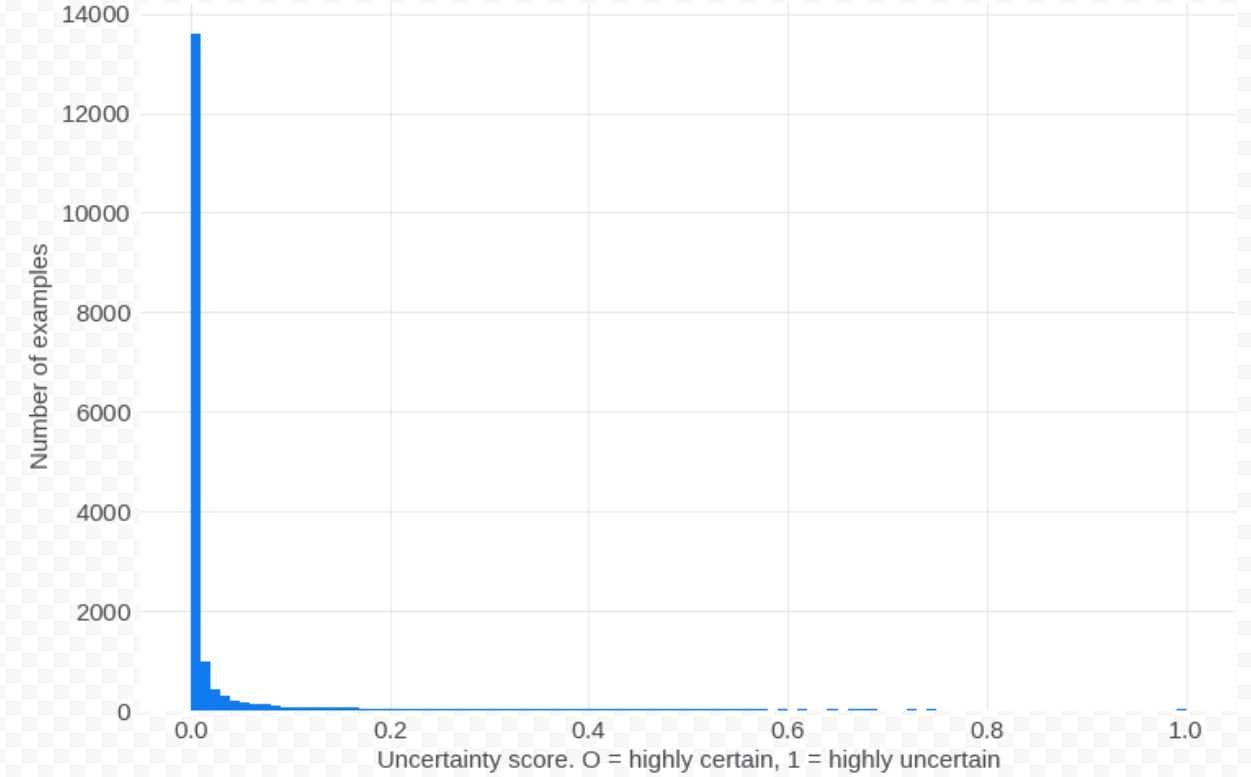}
  \caption{Distribution of uncertainty scores on 20,000 crowd-sourced test examples.}
  \label{fig:uncertainty_overall}
\end{figure}

For our offline evaluation, we computed the simple uncertainty score of the NLU model on 20,000 crowd-sourced test examples. As shown in Figure \ref{fig:uncertainty_overall}, there was a sharp peak of low uncertainty scores. This is not surprising given that NLU model's accuracy on this test set was also high, $>90\%$,.

Next, we observed how the histogram changes if we only looked at misclassified samples. This allowed us to get a sense of how correlated the uncertainty metric is with misclassified points. A perfect histogram would have a sharp peak at 1.0 and very little presence at 0. 

\begin{figure*}[t]
  \centering
  \includegraphics[scale=.3]{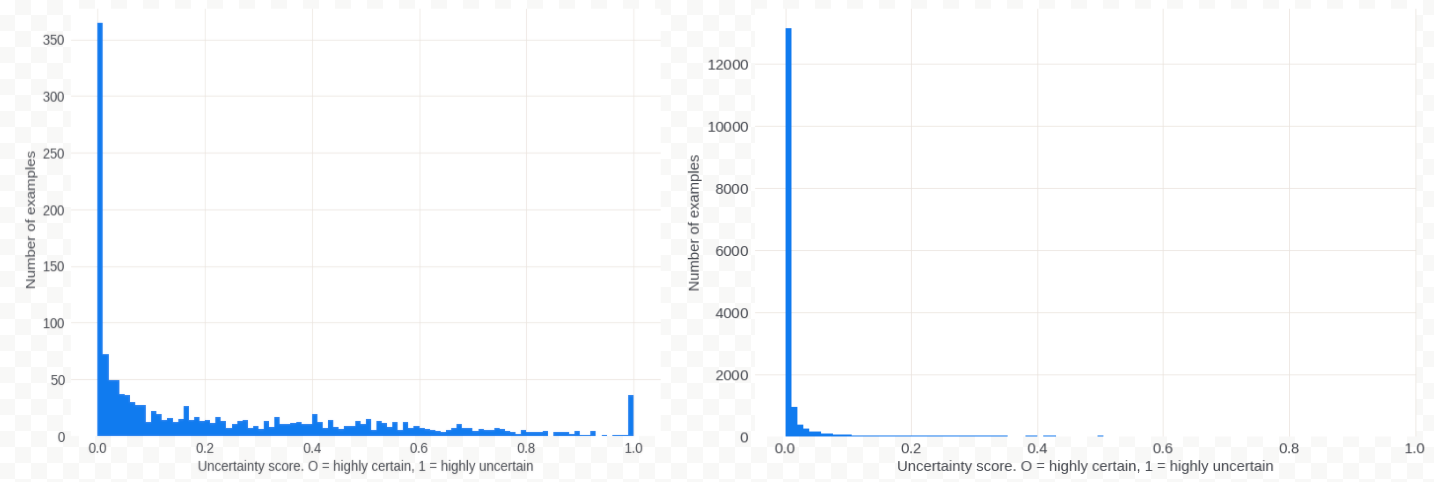}
  \caption{Distribution of uncertainty scores on 20,000 crowd-sourced test examples. The histogram on the left only contains misclassified samples, and the histogram on the right only contains correctly classified samples.}
  \label{fig:uncertainty_comparison}
\end{figure*}

As shown in Figure \ref{fig:uncertainty_comparison} (left), we found that the model is unfortunately highly certain (has low uncertainty) about most of the misclassified samples. However, compared to Figure \ref{fig:uncertainty_comparison} (right), which only contains correctly classified samples, there is a much greater tendency for misclassified samples to have high uncertainty scores than there is for correctly classified samples to have high uncertainty scores.

Thus, despite the NLU model being poorly calibrated -- in fact, only 27\% of the misclassified samples have an uncertainty higher than 0.5 -- we found that \textbf{for sufficiently small values of $k$ i.e. $k << n$, sampling the top $k$ utterances by uncertainty score is likely to yield mostly misclassifications i.e., NLU bugs}.

\subsubsection{Online Evaluation}
For the online evaluation, we created a candidate pool of 30 days of production data from September 2021. The size \textit{n} of the candidate pool was approximately one million requests.
From this candidate set, we independently applied two sampling strategies. The first sampling strategy was random selection. The second sampling strategy was simple uncertainty selection, which we described in Section \ref{section:simple_uncertainty}. We set $k=10,000$. The number of utterances we wanted to sample was an order of 100 times smaller than the candidate pool, i.e., $k << n$.

For each of these sampled sets, we then counted the number of sampled utterances that were not part of a successful \textit{task}. A dialog \textit{task} is a higher-level abstraction than an intent. It may contain multiple back-and-forth interactions between the user and the voice assistant. Refer to \cite{bing} for more details on tasks and computing task success rate. 

We did not grade the task success manually, but rather used an automated metric for estimating task success that correlates closely with the true task success rate. This automated metric checks whether the final NLG dialog act was to \texttt{inform} the user by providing a response (as opposed to replying with an error message). While this may over-estimate true task success, it serves as a good proxy.

Returning to our comparison of the sampling strategies, we found that uncertainty sampling yielded \textbf{four times more task \textit{failures}} than random sampling. The raw counts are given in Table \ref{table:task_failures}. 

\begin{table*}[t]
\centering
\caption{For each sampling strategy, we counted the number of sampled utterances that were part of a task failure.}
\label{table:task_failures}
\vskip 0.15in
\begin{small}
\begin{sc}
\begin{tabular}{lcccr}
\toprule
Sampling Strategy & \# of Sampled Rows & \# of Task Failures\\
\midrule
Random                & 10K    & 1394 \\
Simple Uncertainty    & 10K    & 5620 \\
\bottomrule
\end{tabular}
\end{sc}
\end{small}
\vskip -0.1in
\end{table*}

Note that an NLU prediction for a given utterance can be correct, even when there is task failure; inversely, an NLU prediction for a given utterance can be incorrect, despite a task success. So while these results suggest that simple uncertainty will find more NLU bugs than random sampling in an online setting, it does not guarantee it.

\subsection{Data Augmentation}
\label{section:data_aug_eval}

For our final experiments, we tested the effectiveness of automated data augmentation at correcting bugs deemed to have the \textit{low training data} error attribution. We compared models trained with three different data augmentation strategies:

\begin{itemize}
    \item \textbf{Baseline} An NLU model with no data augmentation applied.
    \item \textbf{Data Augmentation (Exact Match)} For each bug, we added the exact utterance to the training data. For example, for the bug, \textit{``Please call mom,"} we also added, \textit{``Please call mom"} to training. Each training utterance was given a weight of 5 (copied 5 times).
    \item \textbf{Data Augmentation (Templated)}  For each bug, we generated 0 to 5 templates, each of which was expanded up to 10 times. For example, for the bug, \textit{``Please call mom,"} we created the template \texttt{Please call <SL:CONTACT>}. This would then be expanded to \textit{``Please call dad,"} \textit{``Please call Jill,"} etc. and added to training.
\end{itemize}

We used a set of 1200 unique bug utterances, sampled from production data, as seeds for data generation. We then evaluated the performance of the augmented models against the original bug utterances, as well as our validation and test set (which were randomly sampled from production data), and finally a crowd-sourced test set. The results of the augmentation are shown in Table \ref{table:data_aug}.

\begin{table*}[t]
\centering
\caption{We show the improvement that automated data augmentation techniques have on various test sets.}
\label{table:data_aug}
\vskip 0.15in
\begin{small}
\begin{sc}
\begin{tabular}{lcccr}
\toprule
 & Unique Bugs & Production Validation  & Production Test & Crowd-sourced Test \\
 & ($n=1207$) & ($n=66253$) & ($n=32854$) & ($n=95107$) \\
\midrule
Baseline & 5.5\% & 96.1\% & 93.9\% & 90.9\% \\
Exact Match & 95.8\% & 98.0\% & 95.4\% & 90.4\% \\
Templated & 70.6\% & 97.5\% & 95.0\% & 90.0\% \\
\bottomrule
$\Delta$ Templated & +65.1\% & +1.4\% & +1.1\% & -0.9\% \\
\end{tabular}
\end{sc}
\end{small}
\vskip -0.1in
\end{table*}

Both the exact match and templated data augmentation strategies led to significant improvements on the data sets shown. \textbf{Using templated data augmentation, we were able to automatically correct 65\%, i.e., 789 out of 1207, of identified bugs.} This was an order of magnitude higher than the number of bugs we are able to correct in the prior three months using manual data augmentation. 

Interestingly, exact match performed even better than templated data augmentation, which went against our hypothesis: we had expected templated augmentation to lead to a better model, since the additional variety in the training data could improve generalization. We believe the reason the performance of templated data was slightly worse was due to noise in the extracted templates and slot substitutions, which we discuss further in Section \ref{section:qualitative}.

Finally, another point worth mentioning is that while performance on the data sets sampled from production traffic increase, performance on the crowd-sourced data set actually dropped slightly. Digging into this, we found the reason for the drop was because production data and crowd-sourced data sometimes had the same or similar utterances annotated slightly differently. Apart from the labeling discrepancy, there could also be a genuine domain shift between our production and crowd-sourced data.
    
\subsection{Qualitative Feedback}
\label{section:qualitative}
In addition to the evaluations above, we also received qualitative feedback from linguists, who were the primary users of the AutoNLU user interface. We share a few highlights of the constructive feedback:

\textbf{Include humans-in-the-loop while generating data.} Although the original goal of AutoNLU was to automate as much of the quality improvement process as possible, we found it was beneficial, and in many cases necessary, to have a human i.e, a developer or linguist, review the generated data prior to landing it in the training set. Templated data generation can produce nonsensical utterances and sometimes even incorrect annotations. Until better data generation strategies are developed, a human is needed to catch and remove incorrect examples.

\textbf{Make it easy to update annotations across data sets.} Another suggestion we received was to make it easier for linguists to update incorrect annotations that are spread across multiple data sets. A change in the annotation of an utterance in one data set should propagate to others.

\textbf{Track bug fixes landed over time.} Finally, linguists requested the ability to track bugs that have been fixed in the past. This empowers linguists to measure the impact of their work, and enables tracking of when a bug recurs again.

\section{Related Work}
Previous work applying active learning to NLU focuses on improving performance on a fixed test set with fewer training examples \cite{overnight_al, amazon_al, apple_al}; instead, we propose framing active learning as a problem of continuous bug detection.

Valuable libraries and tools for model interpretability and error analysis exist, some specifically tailored for NLP \cite{captum, lit}. However, the debugging information provided by these tools is quite granular. The error classifications provided by AutoNLU are intended to be higher-level and map to a concrete next-action that a linguist or developer can take to resolve a production bug. 
 
On the error correction side, there has been recent work in programmatic labeling \cite{snorkel} and frameworks for adversarial data collection and generation for NLP \cite{dynabench, textattack}. However, these stand-alone libraries are not tied to tools for model and training data analysis. AutoNLU, in contrast, attempts to bring error attribution and correction together. 
 
Finally, in industry, there has recently been work by HuggingFace called \href{https://huggingface.co/autonlp}{AutoNLP} and work by DeepOpinion also called \href{https://deepopinion.ai/autonlu}{AutoNLU}. Despite their similarity in name, these libraries have a different purpose than the AutoNLU described here: they allow users to create new NLP or NLU models in a no- or low-code way. The AutoNLU we present here has the entirely different goal of automating the quality improvement process of already-built NLU models.

\section{Conclusion}
\paragraph{Key Takeaways} We presented a system called AutoNLU, which aims to automate three steps of the NLU quality improvement process: error detection, attribution, and correction. 

We showed that error detection can be done with active learning methods. A simple least-confidence uncertainty baseline can be surprisingly effective at finding failed tasks, even with an uncalibrated model, as long as the number of utterances to be sampled $k$ is significantly smaller than the number of candidates $n$ (in our experiments, $n$ was 100 times larger than $k$). In an online test, we found four times as many failed tasks with this method as we did with random sampling.

In addition, we presented some heuristics for attributing the cause of a model error and strategies for fixing these errors. We showed that templated data generation, where entities inside of slots are substituted, is an effective and scalable method for fixing NLU bugs. However, a human-review step may still be required to inspect the generated data. 

Using this method, we were able to auto-fix nearly 800 bugs from production traffic, a scale ten-fold of what we were able to achieve when doing data augmentation by hand. Despite these gains, we acknowledge that there are limitations of automated data generation, such as generating nonsensical utterances, that will need to be addressed before fully removing humans from the loop.

\paragraph{Future Directions} 
We believe there are many interesting avenues to improve AutoNLU. On the active learning front, we hope to test both new intrinsic and extrinsic sampling strategies to see if they yield an even higher rate of NLU bugs and failed tasks. \cite{shrey} showed that an extrinsic, SVM model can accurately predict when a Transformer model will make a mistake. This SVM could be trained to incorporate features only available at runtime, such as whether the user had to repeat their request. Another aspect to think about is the quality of bugs selected to fix. BatchBALD \cite{batchbald} may help eliminate sampling of redundant bugs and increase their diversity, for example. 

Error attribution could be improved by providing more fine-grained classifications that rely less on heuristics on the training data but rather a more nuanced view of the model's uncertainty. For example, being able to distinguish between \textit{epistemic} uncertainty (uncertainty due to a fundamental lack of knowledge) and \textit{aleatoric} uncertainty (uncertainty due to probabilities being distributed across multiple plausible outcomes) \cite{uncertainty_types} would also enable us to better decide if a bug is caused because of a lack of training data or labeling conflicts.

Finally, on the data augmentation front, we are excited about the potential of using language models to generate data \cite{gpt3}, which may produce more natural new training examples than the data generated by a rule-based template system. 

\section{Acknowledgements}
We would like to thank Geoffrey Goh for building the data augmentation library described in Section \ref{section:data_aug_tool}. We would also like to thank Rongrong Qiu, George Guliashvili, TJ Singh, Cherry Wang, Shicong Zhao, and Nancy Li for their work and guidance on building the interface shown in Figure \ref{fig:pretty_interface}.

We would like to thank Shrey Desai, Michael Marlen, Maria Sumner, John Bourassa, Julien Odent, Bing Liu, Ousmane Dia, Ted Wild, Jeffrey Kyle Chiu, Zoe Papakipos, Alexander Zotov, Sonal Gupta, Anuj Kumar, Jinsong Yu, and Luna Dong for their invaluable early discussions, feedback, and support of this work.

Finally, we would also like to thank Justin Rill, Brian Moran, Claire Lesage, Tammy Stark, Caitlin Lohman, Jeremy Kahn, and Safiyyah Saleem for being patient dogfooders of the prototype and providing an invaluable linguist perspective on this work.

\bibliography{autonlu}

\begin{thebibliography}{27}
\providecommand{\natexlab}[1]{#1}
\providecommand{\url}[1]{\texttt{#1}}
\expandafter\ifx\csname urlstyle\endcsname\relax
  \providecommand{\doi}[1]{doi: #1}\else
  \providecommand{\doi}{doi: \begingroup \urlstyle{rm}\Url}\fi

\bibitem[Breck et~al.(2019)Breck, Polyzotis, Roy, Whang, and
  Zinkevich]{data_validation}
Breck, E., Polyzotis, N., Roy, S., Whang, S.~E., and Zinkevich, M.~A.
\newblock Data validation for machine learning.
\newblock In \emph{MLSys}, 2019.

\bibitem[Brown et~al.(2020)Brown, Mann, Ryder, Subbiah, Kaplan, Dhariwal,
  Neelakantan, Shyam, Sastry, Askell, Agarwal, Herbert-Voss, Krueger, Henighan,
  Child, Ramesh, Ziegler, Wu, Winter, Hesse, Chen, Sigler, Litwin, Gray, Chess,
  Clark, Berner, McCandlish, Radford, Sutskever, and Amodei]{gpt3}
Brown, T.~B., Mann, B., Ryder, N., Subbiah, M., Kaplan, J., Dhariwal, P.,
  Neelakantan, A., Shyam, P., Sastry, G., Askell, A., Agarwal, S.,
  Herbert-Voss, A., Krueger, G., Henighan, T.~J., Child, R., Ramesh, A.,
  Ziegler, D.~M., Wu, J., Winter, C., Hesse, C., Chen, M., Sigler, E., Litwin,
  M., Gray, S., Chess, B., Clark, J., Berner, C., McCandlish, S., Radford, A.,
  Sutskever, I., and Amodei, D.
\newblock Language models are few-shot learners.
\newblock \emph{ArXiv}, abs/2005.14165, 2020.

\bibitem[Chen et~al.(2019{\natexlab{a}})Chen, Zhuo, and Wang]{bert}
Chen, Q., Zhuo, Z., and Wang, W.
\newblock Bert for joint intent classification and slot filling.
\newblock \emph{ArXiv}, abs/1902.10909, 2019{\natexlab{a}}.

\bibitem[Chen et~al.(2019{\natexlab{b}})Chen, Sagar, Kao, Li, Klein, Pulman,
  Garg, and Williams]{apple_al}
Chen, X.~C., Sagar, A., Kao, J.~T., Li, T.~Y., Klein, C., Pulman, S.~G., Garg,
  A., and Williams, J.
\newblock Active learning for domain classification in a commercial spoken
  personal assistant.
\newblock In \emph{INTERSPEECH}, 2019{\natexlab{b}}.

\bibitem[Desai \& Aly(2021)Desai and Aly]{shrey}
Desai, S. and Aly, A.
\newblock Diagnosing transformers in task-oriented semantic parsing.
\newblock \emph{ArXiv}, abs/2105.13496, 2021.

\bibitem[Duong et~al.(2018)Duong, Afshar, Estival, Pink, Cohen, and
  Johnson]{overnight_al}
Duong, L., Afshar, H., Estival, D., Pink, G., Cohen, P., and Johnson, M.
\newblock Active learning for deep semantic parsing.
\newblock In \emph{Proceedings of the 56th Annual Meeting of the Association
  for Computational Linguistics (Volume 2: Short Papers)}, pp.\  43--48,
  Melbourne, Australia, July 2018. Association for Computational Linguistics.
\newblock \doi{10.18653/v1/P18-2008}.
\newblock URL \url{https://aclanthology.org/P18-2008}.

\bibitem[Gupta et~al.(2018)Gupta, Shah, Mohit, Kumar, and Lewis]{gupta}
Gupta, S., Shah, R., Mohit, M., Kumar, A., and Lewis, M.
\newblock Semantic parsing for task oriented dialog using hierarchical
  representations.
\newblock In \emph{EMNLP}, 2018.

\bibitem[He et~al.(2020)He, Zhang, Lei, Chen, Chen, Alhamadani, Xiao, and
  Lu]{overconfidence}
He, J., Zhang, X., Lei, S., Chen, Z., Chen, F., Alhamadani, A., Xiao, B., and
  Lu, C.-T.
\newblock Towards more accurate uncertainty estimation in text classification.
\newblock In \emph{EMNLP}, 2020.

\bibitem[Huang et~al.(2015)Huang, Xu, and Yu]{bilstm_crf}
Huang, Z., Xu, W., and Yu, K.
\newblock Bidirectional lstm-crf models for sequence tagging.
\newblock \emph{ArXiv}, abs/1508.01991, 2015.

\bibitem[Kiela et~al.(2021)Kiela, Bartolo, Nie, Kaushik, Geiger, Wu, Vidgen,
  Prasad, Singh, Ringshia, Ma, Thrush, Riedel, Waseem, Stenetorp, Jia, Bansal,
  Potts, and Williams]{dynabench}
Kiela, D., Bartolo, M., Nie, Y., Kaushik, D., Geiger, A., Wu, Z., Vidgen, B.,
  Prasad, G., Singh, A., Ringshia, P., Ma, Z., Thrush, T., Riedel, S., Waseem,
  Z., Stenetorp, P., Jia, R., Bansal, M., Potts, C., and Williams, A.
\newblock Dynabench: Rethinking benchmarking in nlp.
\newblock \emph{ArXiv}, abs/2104.14337, 2021.

\bibitem[Kirsch et~al.(2019)Kirsch, van Amersfoort, and Gal]{batchbald}
Kirsch, A., van Amersfoort, J.~R., and Gal, Y.
\newblock Batchbald: Efficient and diverse batch acquisition for deep bayesian
  active learning.
\newblock In \emph{NeurIPS}, 2019.

\bibitem[Kluyver et~al.(2016)Kluyver, Ragan-Kelley, P{\'e}rez, Granger,
  Bussonnier, Frederic, Kelley, Hamrick, Grout, Corlay, Ivanov, Avila, Abdalla,
  Willing, and development team]{jupyter}
Kluyver, T., Ragan-Kelley, B., P{\'e}rez, F., Granger, B., Bussonnier, M.,
  Frederic, J., Kelley, K., Hamrick, J., Grout, J., Corlay, S., Ivanov, P.,
  Avila, D., Abdalla, S., Willing, C., and development team, J.
\newblock Jupyter notebooks - a publishing format for reproducible
  computational workflows.
\newblock In Loizides, F. and Scmidt, B. (eds.), \emph{Positioning and Power in
  Academic Publishing: Players, Agents and Agendas}, pp.\  87--90, Netherlands,
  2016. IOS Press.
\newblock URL \url{https://eprints.soton.ac.uk/403913/}.

\bibitem[Kokhlikyan et~al.(2020)Kokhlikyan, Miglani, Martin, Wang, Alsallakh,
  Reynolds, Melnikov, Kliushkina, Araya, Yan, and Reblitz-Richardson]{captum}
Kokhlikyan, N., Miglani, V., Martin, M., Wang, E., Alsallakh, B., Reynolds, J.,
  Melnikov, A., Kliushkina, N., Araya, C., Yan, S., and Reblitz-Richardson, O.
\newblock Captum: A unified and generic model interpretability library for
  pytorch.
\newblock \emph{ArXiv}, abs/2009.07896, 2020.

\bibitem[Koçak et~al.(2021)Koçak, Ramirez, Erkip, and Shasha]{safe_predict}
Koçak, M.~A., Ramirez, D., Erkip, E., and Shasha, D.
\newblock Safepredict: A meta-algorithm for machine learning that uses refusals
  to guarantee correctness.
\newblock \emph{IEEE Transactions on Pattern Analysis and Machine
  Intelligence}, 43:\penalty0 663--678, 2021.

\bibitem[Liu et~al.(2018)Liu, T{\"u}r, Hakkani-T{\"u}r, Shah, and Heck]{bing}
Liu, B., T{\"u}r, G., Hakkani-T{\"u}r, D.~Z., Shah, P., and Heck, L.
\newblock Dialogue learning with human teaching and feedback in end-to-end
  trainable task-oriented dialogue systems.
\newblock In \emph{NAACL}, 2018.

\bibitem[Liu et~al.(2019)Liu, Ott, Goyal, Du, Joshi, Chen, Levy, Lewis,
  Zettlemoyer, and Stoyanov]{roberta}
Liu, Y., Ott, M., Goyal, N., Du, J., Joshi, M., Chen, D., Levy, O., Lewis, M.,
  Zettlemoyer, L., and Stoyanov, V.
\newblock Roberta: A robustly optimized bert pretraining approach.
\newblock \emph{ArXiv}, abs/1907.11692, 2019.

\bibitem[Mesnil et~al.(2013)Mesnil, He, Deng, and Bengio]{atis}
Mesnil, G., He, X., Deng, L., and Bengio, Y.
\newblock Investigation of recurrent-neural-network architectures and learning
  methods for spoken language understanding.
\newblock In \emph{INTERSPEECH}, 2013.

\bibitem[Monarch(2021)]{hitl}
Monarch, R.
\newblock \emph{Human-in-the-Loop Machine Learning: Active learning and
  annotation for human-centered AI}.
\newblock Manning, 2021.
\newblock ISBN 9781638351030.
\newblock URL \url{https://books.google.com/books?id=bNo2EAAAQBAJ}.

\bibitem[Morris et~al.(2020)Morris, Lifland, Yoo, Grigsby, Jin, and
  Qi]{textattack}
Morris, J.~X., Lifland, E., Yoo, J.~Y., Grigsby, J., Jin, D., and Qi, Y.
\newblock Textattack: A framework for adversarial attacks, data augmentation,
  and adversarial training in nlp.
\newblock In \emph{EMNLP}, 2020.

\bibitem[Ratner et~al.(2017)Ratner, Bach, Ehrenberg, Fries, Wu, and
  R{\'e}]{snorkel}
Ratner, A.~J., Bach, S.~H., Ehrenberg, H.~R., Fries, J.~A., Wu, S., and R{\'e},
  C.
\newblock Snorkel: Rapid training data creation with weak supervision.
\newblock \emph{Proceedings of the VLDB Endowment. International Conference on
  Very Large Data Bases}, 11 3:\penalty0 269--282, 2017.

\bibitem[Reback et~al.(2021)Reback, jbrockmendel, McKinney, den Bossche,
  Augspurger, Cloud, Hawkins, gfyoung, Roeschke, Sinhrks, Klein, Petersen,
  Tratner, She, Ayd, Hoefler, Naveh, Garcia, Schendel, Hayden, Saxton,
  Darbyshire, Shadrach, Gorelli, Li, Jancauskas, McMaster, Zeitlin, Battiston,
  and Seabold]{pandas_software}
Reback, J., jbrockmendel, McKinney, W., den Bossche, J.~V., Augspurger, T.,
  Cloud, P., Hawkins, S., gfyoung, Roeschke, M., Sinhrks, Klein, A., Petersen,
  T., Tratner, J., She, C., Ayd, W., Hoefler, P., Naveh, S., Garcia, M.,
  Schendel, J., Hayden, A., Saxton, D., Darbyshire, J., Shadrach, R., Gorelli,
  M.~E., Li, F., Jancauskas, V., McMaster, A., Zeitlin, M., Battiston, P., and
  Seabold, S.
\newblock pandas-dev/pandas: Pandas 1.3.3, September 2021.
\newblock URL \url{https://doi.org/10.5281/zenodo.5501881}.

\bibitem[Ribeiro et~al.(2020)Ribeiro, Wu, Guestrin, and Singh]{checklist}
Ribeiro, M.~T., Wu, T.~S., Guestrin, C., and Singh, S.
\newblock Beyond accuracy: Behavioral testing of nlp models with checklist.
\newblock In \emph{ACL}, 2020.

\bibitem[Sen \& Yilmaz(2020)Sen and Yilmaz]{amazon_al}
Sen, P. and Yilmaz, E.
\newblock Uncertainty and traffic-aware active learning for semantic parsing.
\newblock In \emph{Proceedings of the First Workshop on Interactive and
  Executable Semantic Parsing}, pp.\  12--17, Online, November 2020.
  Association for Computational Linguistics.
\newblock \doi{10.18653/v1/2020.intexsempar-1.2}.
\newblock URL \url{https://aclanthology.org/2020.intexsempar-1.2}.

\bibitem[Shrivastava et~al.(2021)Shrivastava, Chuang, Babu, Desai, Arora,
  Zotov, and Aly]{span_pointer}
Shrivastava, A., Chuang, P. I.-J., Babu, A., Desai, S., Arora, A., Zotov, A.,
  and Aly, A.
\newblock Span pointer networks for non-autoregressive task-oriented semantic
  parsing.
\newblock \emph{ArXiv}, abs/2104.07275, 2021.

\bibitem[Tenney et~al.(2020)Tenney, Wexler, Bastings, Bolukbasi, Coenen,
  Gehrmann, Jiang, Pushkarna, Radebaugh, Reif, and Yuan]{lit}
Tenney, I., Wexler, J., Bastings, J., Bolukbasi, T., Coenen, A., Gehrmann, S.,
  Jiang, E., Pushkarna, M., Radebaugh, C., Reif, E., and Yuan, A.
\newblock The language interpretability tool: Extensible, interactive
  visualizations and analysis for nlp models.
\newblock In \emph{EMNLP}, 2020.

\bibitem[{W}es {M}c{K}inney(2010)]{pandas_paper}
{W}es {M}c{K}inney.
\newblock {D}ata {S}tructures for {S}tatistical {C}omputing in {P}ython.
\newblock In {S}t\'efan van~der {W}alt and {J}arrod {M}illman (eds.),
  \emph{{P}roceedings of the 9th {P}ython in {S}cience {C}onference}, pp.\  56
  -- 61, 2010.
\newblock \doi{10.25080/Majora-92bf1922-00a}.

\bibitem[Ülkümen et~al.(2016)Ülkümen, Fox, and Malle]{uncertainty_types}
Ülkümen, G., Fox, C., and Malle, B.
\newblock Two dimensions of subjective uncertainty: Clues from natural
  language.
\newblock \emph{Journal of Experimental Psychology: General}, 145, 07 2016.
\newblock \doi{10.1037/xge0000202}.

\end{thebibliography}
\bibliographystyle{mlsys2022}



\end{document}